# Novel Machine Learning Approach for Predicting Poverty using Temperature and Remote Sensing Data in Ethiopia

Om Shah[1], Krti Tallam[2]

**Social Sciences, Sustainability Sciences**
**Machine Learning, Poverty Modeling, Climate Change**

1. Independent Researcher, Lakeside School
2. Department of Ecology, Stanford University, Stanford, CA, 94305

## Abstract

In many developing nations, a lack of poverty data prevents critical humanitarian organizations from responding to large-scale crises. Currently, socio-economic surveys are the only method implemented on a large scale for organizations and researchers to measure and track poverty. However, the inability to collect survey data efficiently and inexpensively leads to significant temporal gaps in poverty data; these gaps severely limit the ability of organizational entities to address poverty at its root cause. We propose a transfer learning model based on surface temperature change and remote sensing data to extract features useful for predicting poverty rates. Machine learning, supported by data sources of poverty indicators, has the potential to estimate poverty rates accurately and within strict time constraints. Higher temperatures, as a result of climate change, have caused numerous agricultural obstacles, socio-economic issues, and environmental disruptions, trapping families in developing countries in cycles of poverty. To find patterns of poverty relating to temperature that have the highest influence on spatial poverty rates, we use remote sensing data. The two-step transfer model predicts the temperature delta from high resolution satellite imagery and then extracts image features useful for predicting poverty. The resulting model achieved 80% accuracy on temperature prediction. This method takes advantage of abundant satellite and temperature data to measure poverty in a manner comparable to the existing survey methods and exceeds similar models of poverty prediction.

## Significance Statement

According to the United Nations' Sustainable Development Goals, humanity's first and foremost goal is the elimination of worldwide poverty. Accurate, frequent poverty data is necessary for humanitarian organizations and policy makers to address poverty, especially in developing countries where a lack of poverty data has contributed to the uneven allocation of stimulus and resources. Without correctly directing resources to the impoverished, we cannot uplift communities out of cycles of poverty and successfully eliminate economic status as an obstacle to humanity's long-term success. The novel method to predict poverty discussed in this paper is essential to the critical need for poverty data in developing countries.

## Introduction

There is a pressing need for inexpensive, abundant poverty data in developing nations. The United Nations' (UN) World That Counts publication reports the repercussions of lacking data are serious, ranging from a neglect of human rights to rapid environmental degradation. Sub-annual data is needed for countries to combat poverty through policy decisions, NGO funding, and effective disaster response. While sufficient poverty data is available for wealthy countries, through a combination of active and passive data sources, to properly accommodate humanitarian services, many developing countries face data shortages that limit their ability to serve at-risk communities [1]. The current prevailing data collection method, surveys, are time-consuming and expensive due to a variety of factors. Inefficient questionnaires decrease the quality of data collected while driving costs up. Furthermore, substantial portions of the population live in inaccessible rural areas making survey logistics difficult to plan [2]. Governments with strict budgeting requirements may not be willing to spend funds on employing survey personnel, or local leadership could harbor dissent towards data collection associated with a particular government or organization [3]. There is no consistent method for high quality data collection as any number of factors could impact proper execution. However, recent efforts focus on the use of machine learning to create poverty prediction solutions that utilize publicly available datasets and increase efficiency. Stanford University's Ermon Lab produced a machine learning transfer model that utilized nightlights as a proxy for urbanization and satellite imagery based image classification to support a poverty prediction task in multiple African countries. We extend the remote sensing based poverty mapping work researched by the Ermon Lab to examine the causal temperature-consumption relationship [4].

Causal relationships between consumption and consumption-determinators such as temperature are often viewed as downstream tasks [5]. In most cases, consumption-determinators are used to inform policy decisions after a deep learning model has been trained to predict consumption. It can be useful, however, to exploit the causal relationship directly in model training rather than as a corroboration of findings. Examining a singular causal relationship doesn't convey the full picture of poverty rates, but allows for segmenting determinants by their ability to explain variation in poverty rates. For example, organizations with climate focuses can direct resources towards areas most vulnerable to climate change, as seen through a temperature-consumption model. Monetary policy can be targeted towards economic relief for communities bearing the brunt of rising temperatures.

An opportunity exists to solve the poverty data void through novel methods. The United Nations wrote in their 2020 World Social report that urbanization is linked to poverty: most of the world population in poverty live in rural areas. Migration from rural areas to urban areas can result in a reduction in poverty in many cases. However, the transfer of people between rural and urban areas is a long term effect and thus can only be measured through long term analysis [6]. We therefore look at a different cause of poverty linked to variance in regional poverty rates. For populations living in agricultural communities, climate change poses a greater threat. Rising global temperatures are drastically reducing crop harvest output in regions such as sub-saharan Africa where poverty rates exceed 40%, impacting families who rely on agricultural growth to escape poverty [7, 8]. In addition to crop failures, high temperatures have also caused the evaporation of water sources and placed a financial burden on families to improve their roofing material [9, 10]. What otherwise may have been a temporary period of low income and crop yields has transformed into long term financial distress.

Global average temperature data can statistically represent the progress of climate change in the context of deep learning. Not only is temperature data abundant, but advances in satellite resolution have produced highly accurate climate datasets. However, by directly relating temperature data to poverty, we cannot assign weights to specific patterns between temperature and poverty. For example, visual patterns such as evaporating water sources or crop color combined with a high temperature correlation may indicate higher poverty rates than roof material. Remote sensing data is a potential data source for finding the strongest spatial links between temperature and poverty because satellite imagery can be collected instantaneously and inexpensively. Satellite data can measure a variety of natural and human developments such as water sources, roofing material, and crops, but also micro-features that are only a few pixels wide. Recent developments in machine learning tools and technology have increased computing efficiency for analyzing sophisticated forms of data such as satellite imagery. Machine learning models can be trained through indicators of climate change - namely temperature - to extract features from satellite imagery and produce meaningful poverty predictions.

We propose using a transfer learning model trained on temperature data to extract features from satellite images and predict poverty in Ethiopia. While transfer learning is traditionally used to compensate for a lack of training data in a target problem, we utilize it as a way to introduce spatial observations of the indirect temperature-consumption relationship. The first step will predict temperature from satellite images to learn features useful in predicting poverty. The second step will extract a feature vector from the temperature model and use those features to linearly predict consumption, an economic indicator of poverty.

Our three primary data sources are the satellite images, temperature data, and survey data. Satellite images are downloaded from the Google Static Maps API at a 16 factor zoom resolution. Temperature data is aggregated from WorldClim, a large scale climate database, at a 30 degree arc second resolution (approximately 1km by 1km). In this paper, average temperature refers to the average delta temperature between 1980 and 2016 for Ethiopia. Consumption per capita data is collected by the Living Standards Measurements Study (LSMS) 2016 socio-economic survey for Ethiopia.

The first step of the transfer learning model predicts temperature from satellite images. A pretrained convolutional neural network (CNN) model predicts temperature from satellite images and outputs a trained model from which features can be extracted. The CNN model is pre-trained so it can skip over unnecessary training that learns basic image patterns recognized in most image recognition models [11]. The first layers of any CNN model will almost always find relationships in images such as edges and corners. To save time and energy on the computing device, we avoid retraining for elementary patterns.

Feature extraction extracts the most useful features from each image and compiles a feature vector for the trained model [12]. In most image classification cases, feature extraction must be used because the sheer size of features from a large dataset of images can lead to unpredictable issues in later stages. Without feature extraction, the second part of the model trains on a noisy feature set and therefore produces a less accurate result.

The second step of the model involves training a ridge regression model on the extracted feature set. Ridge regression, a form of linear regression, is used to fit the consumption per capita values to the feature set. Unlike standard linear regression, a ridge model adds a bias to the hypothesis function to reduce overfitting and create a function better suited for prediction [13]. This proves useful if we wish to implement the model in different geographic locations or with contrasting demographic data.

We test the accuracy of our model using a five-fold cross validation model. With a standard holdout validation system, a model is tested using a limited set of test data which may introduce bias. The alternative method is a k-fold cross validation. Cross validation is a process of testing a model iteratively based on subsections of training data to ensure limited bias [14]. We limit bias because the model may have performed differently based on distinct geographic data trends within Ethiopia. The correlation of actual and predicted consumption values can be measured with an $R^2$ value, also known as the coefficient of determination. The $R^2$ value, a common metric in the machine learning and statistics fields, will inform us of the percentage of poverty rates in Ethiopia explained by the temperature model.

Temperature data and satellite images from Ethiopia can be transformed into highly accurate poverty predictions through machine learning pattern analysis techniques.

## Methods

The transfer learning model consists of two parts: predicting temperature using satellite images and predicting consumption from specific features of satellite images trained on temperature. Temperature data became a potential variable due to its relationship to poverty as an indicator of climate change. Only a specific range of temperatures - especially in developing countries - can support profitable crop yields, provide an environment with adequate water resources, and allow for unobstructed living [8, 9, 10]. In many places, the window of viable agricultural temperature is quickly closing in and water sources are rapidly depleting. We use temperature as a proxy for climate change and its impacts on poverty. Global average temperatures have been tracked by satellites for decades and their accuracy is getting more precise by year. The temperature data used in this paper has a spatial resolution of 30 arc seconds which equates to approximately a 1km by 1km resolution.

While temperature data can be computed for any latitude and longitude in the world, consumption data can only be collected by various surveys that occur every couple of years on a household level. A survey that computes consumption per capita is the Living Standards Measurement Study (LSMS) that is conducted in numerous countries across the globe. Their initiative is linked with the World Bank making the dataset a reliable source of information. The data used in this paper is from the Socioeconomic Survey 2015-2016 in Ethiopia. Asset wealth was the targeted economic measure in other studies such as the one conducted by Yeh et al., where urbanization proxies such as nightlights are indicative of a net-positive collection of material resources by a population - roads, cars, building sizes. We chose to use consumption because the temperature-consumption relationship impacts the ability of households to achieve a net-positive income. Another benefit of using consumption data to test our model against is that we not only get the specific consumption values but also locations where those consumption values were collected. This is helpful because we don't have to find a separate list of latitudes and longitudes for where those consumption values may have been surveyed. Consumption is used as total consumption per capita, a popular metric of poverty in academia and industry applications [15]. We derive consumption per capita to normalize the consumption per individual which is a constant value in all locations instead of per household for which size can fluctuate. We calculate this by taking the total annual consumption of each household and dividing the value by the total number of household members.

Procuring temperature data involves manipulating GeoTIFF data types and superimposing them onto the consumption data. WorldClim is an online repository of climate data from which we used average global temperature in degrees Celsius. A GeoTIFF file consists of an underlying map with embedded metadata [16]. TIFF files cover an entire area instead of specific locations like the LSMS survey. Due to this, any location derived from the LSMS data can be found in the TIFF file with its corresponding temperature. To further normalize the temperature data going into the model, a 10km by 10km bounding box around each location is created. We take the average temperature for this 100km^2 area instead of a single point at the latitude longitude provided by the survey data. For each location, in the form of latitude and longitude, from the LSMS survey, a consumption per capita value and average temperature value is assigned. This aggregated data is used to create specific download locations for the satellite images. To reciprocate taking the average temperature of a 10km by 10km bounding box, multiple images from each bounding box are taken instead of one image per box. This makes sure that there are enough training images for the model to predict temperature from validation images and subsequently predict consumption.

Before downloading the images, a Gaussian mixture model is used to split the temperature data into three bins. Data binning is a pre-processing technique where data is placed into separate bins depending on their relation to measures of central tendency [17]. Each bin is represented by a central value of the data it covers, such as the mean. The mixture model segments any input data into n groups, similar to a k-means model. Each image location, for which an image will be downloaded, is assigned to a temperature bin based on its average temperature. This step will simplify the model so it only has to predict the temperature bin of a model instead of its exact temperature. Another benefit of using temperature bins is that they will remove any anomalies that may arise from the data due to isolated geographic events. This allows the long term temperature increase induced by climate change to shine through any noise. Alongside the original temperature and consumption data, the assigned temperature bin from the gaussian mixture model is added to the set of data points for each image.

To download the images, we used the Google Static Image API. The only parameters needed by the API are latitude, longitude, and the zoom for the images. The API doesn't allow queries for specific timestamps, so the images downloaded were the most recent images taken by satellites. These images are likely <1 year old. We assume that model performance would increase if 2016 satellite images were downloaded. Once this step is complete, all images are downloaded to their respective locations on the hardware.

The model used for predicting is a VGG-11 model pre-trained on ImageNet. The VGG CNN model achieved high accuracy at the 2014 ILSVRC and thus proved to be a viable model for predicting poverty [18]. We chose the VGG-11 model over newer architectures such as ResNet and VGG-19 as the experimental goal was to simply find notable features for classification into three bins. A more advanced model would not return an appropriate performance boost for the additional computing power needed. The numeral after VGG represents the number of weighted layers that each image will be passed through; in VGG-11, there are 11 weighted layers through which the satellite images will be passed. Any VGG model takes in an input image of 224 by 224 pixels. The first eight layers of the model are a combination of convolution and max pooling. These make up the convolutional layers of the VGG-11 model. Each convolutional layer throughout the entire course of the model uses 3x3 filters, the smallest group of pixels within the image, to train. Every layer, the number of filters doubles to increase the model's exposure to the image. The last three layers are dedicated to a fully connected neural network and a softmax activation function. The first two fully connected layers have 4096 channels each while the third layer has 1000 channels. The model outputted a temperature bin prediction for each image. We used a 20 epoch distribution for training the model. Out of the 20 epochs, ten will be used to train the last few layers while the rest will be used to train the entire model. We didn't have to dedicate all the layers because the first layers are pretrained. This highlights the benefit of using a pretrained model to efficiently utilize compute and memory. After the 20 epochs of training are complete, the next step is to extract features from the trained model. The model was trained on a batch size of 8.

Additionally, we introduced augmentation and normalization functions to the model to vary input data. Augmentation involves performing minor edits such as flipping, reflecting, or color grading on images [19]. This increases the number of unique training data that the model receives on which it can train. Normalization takes a set of data and fits it into a given range. We must use normalization because larger values intrinsically have a greater impact on the model than smaller values. Normalization ensures there is no imbalanced influence of the training data on the model.

Feature extraction is a critical step to analyzing the output of the trained model. The VGG-11 CNN model generates thousands of features from satellite images it finds important for predicting temperature. However, many of these features are discardable and limit the efficiency of the predictions [12]. To select what features we want to use, we must pass the model through a feature extraction process. The model used to extract features is a sequential classifier with only the first four layers. From the first layer, 25,088 features are taken as an input and by the last layer, 4096 features are outputted for each image. After the sequential layers, all remaining features are aggregated and compiled into a single feature vector for predicting consumption.

Predicting consumption from the extracted features is conducted through ridge regression. We avoid optimizing the ridge model parameters and architecture as a performance boost to the training is unnecessary. The experiment focuses on measuring the linear relationship between temperature and poverty that inspired the use of ridge regression rather than pushing the boundaries on performance. Ridge regression models add penalties, called a bias, to the hypothesis to avoid overfitting the data and its outliers. By adding penalties, the model performs with greater accuracy on new data. This is ideal because there are still locations where poverty has not been predicted with the model. Ridge models consist of a cost function to calculate the mean squared error differences between predicted and actual output values for each training example. The squared magnitude of all coefficients is added to the overall cost function to skew any produced coefficients toward a higher cost [13]. It is important to note that the square magnitude is first multiplied by the optimal learning rate, alpha, before being added to the overall cost. This allows for faster fitting to the data. For each cluster of training images, multiple images from a similar location in Ethiopia, there is a 4096-feature vector. The feature vectors of all clusters are scaled to create a consistent set of training variables. The training variables are fed into a ridge regression model instantiated by pre-built machine learning libraries, which in this case is scikit-learn. The best alpha for the ridge model is found by iterating over a possible set of hard coded alphas and feeding them into the ridge model. The best alpha is the value which produces the highest $R^2$ value.

Finding the accuracy of predicting consumption to track poverty is the final piece of the transfer learning model. We use a five fold cross validation model to test the performance of the trained ridge model; the five fold cross validation model used is a variant of industry standard k-fold cross validation. Instead of validating the model on a part of the data reserved solely for validation purposes, the data is segmented into k chunks. Holdout is performed on each chunk iteratively and the mean squared error is averaged over each fold [13]. With k-fold cross validation, each piece of training data will be trained on k-1 times and validated with k times. The primary reason for using k-fold cross validation is to limit any selective bias that may occur when separating the training data from validation data. We used a five-fold model because it balances out computational time with the least bias from the training data. Finally, the results of the five fold cross validation are graphed out.

# Results

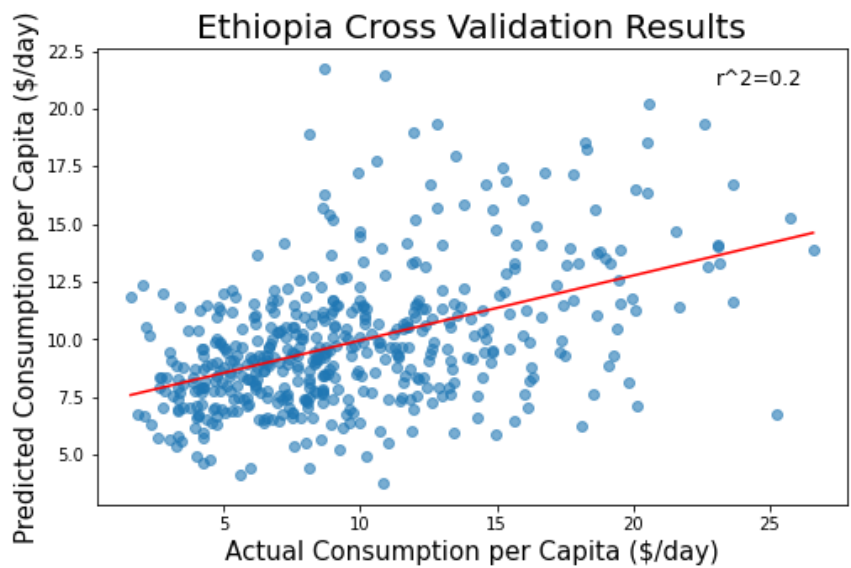

**Figure 1.** The results of the cross validation model are shown with the actual consumption per values plotted on the x-axis and the predicted consumption per capita values plotted on the y-axis. The R^2 value represents the amount of variance of the dependent variable that can be explained by the independent variable.

The trained model displayed an accuracy of 80% over all iterations conducted in 20 epochs for predicting temperature. During the five fold cross validation phase, the model achieved an R^2 score of 0.20. The R^2 score was calculated by dividing the residual sum of squares by the total sum of squares and subtracting the resulting value from 1. As seen by the graph, the predicted function should be as close to the equation f(x)=x as possible; this shows that every predicted consumption value is as close as possible to the actual consumption value.

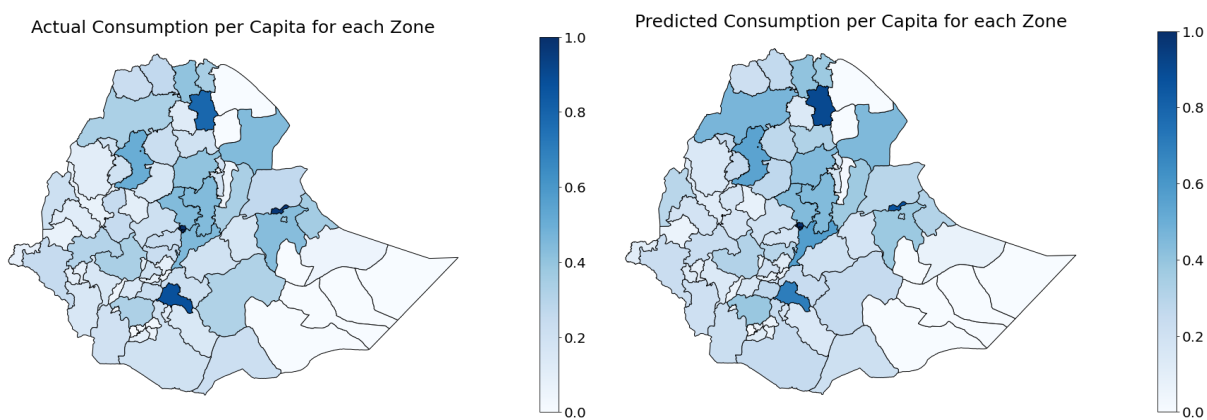

**Figure 2.** The scale indicates normalized total consumption per capita for each administrative zone in Ethiopia for actual versus predicted consumption (left and right respectively) at LSMS survey locations. Model performance in administrative zones is important for organizations and committees to pinpoint specific regions where poverty needs to be addressed.

The temperature model correctly identifies trends throughout Ethiopia where poverty is high. As a whole, the model matches the actual consumption data in spatial poverty mapping: the northern regions have higher poverty rates than southern regions with the exceptions of outliers.

The temperature model can be validated by established indicators of poverty such as NDVI [20] and population data. We ran two additional models using these variables to compare to the accuracy of the temperature model. Normalized difference vegetation index (NDVI), is calculated by measuring the difference between near-infrared light and red light. The data values have a range of -1 to 1. NDVI was used because it can serve as an indicator of land health, soil fertility, and rainfall. Due to this, it proves to be an indicator of poverty most similar to temperature data because both variables consider agriculture as a primary source of sustenance of families.

NDVI data was collected from the Google Earth Engine software in the form of a TIFF file. The data was imposed onto individual download locations and matched with consumption data in a similar process as detailed above. NDVI ran through the complete pipeline and achieved an accuracy of 71.07% for the temperature prediction task with an $R^2$ score of 0.04 for predicting consumption. Like NDVI, population data was also collected using the Google Earth Engine in the TIFF format. Training on population data takes a different approach to predicting poverty. Instead of looking for the impacts of climate change, it's an indicator of urbanization and migration. After running through 20 epochs, the model received an accuracy of 70.05% for the temperature prediction task with an $R^2$ score of 0.05 for predicting consumption. The results of both of these models demonstrates the validity of temperature data and its expertise in predicting poverty.

## Conclusion

While NDVI and population data offer important insights into the performance of alternative models, average temperature still reigns in terms of model performance and its global links to poverty. NDVI has real world ties to crop output and climate change, but it may lead to incorrect model assumptions with a vastly different geographic setting. Despite being located in a desert, with little to no vegetation, many North-African and developed Middle Eastern countries have a poverty rate at or lower than 30%. The visible geographic differences are too literal for accurately predicting poverty across a variety of environments. With population, another issue may arise. Populations for individual regions are primarily calculated through state-run surveys that happen every decade or so. In the off-years when surveys are not being conducted, large migratory patterns could arise and disappear. Another problem is that timely and accurate population data from any given year cannot be fed into the model. This nullifies the purpose of switching to a remote sensing based model where images are available readily and through a variety of data sources.

In comparison to Yeh et al, our model solely focuses on the transfer learning method of poverty prediction. The transfer learning approach was intuitively thought to bolster the inherent benefits of using satellite imagery over a direct linear regression prediction. In Yeh et al, transfer learning was found to perform worse compared to a complete end-to-end model based on nightlights. The increased performance of an end-to-end model in the study points to the transfer learning discarding important spatial features critical to predicting poverty. In a future study, we wish to implement this approach to analyze the benefits of directly measuring poverty without a transfer step.

The results from the temperature model therefore represent the best version of poverty tracking conducted within this study and on track to achieving the results of nighttime imagery models. The model was applied to predict Ethiopia's poverty, but the model has the ability to be used in any geographic location due to the ubiquitous access of temperature data. Temperature data does have its own caveats: notably, climatic anomalies can impact short term poverty tracking. Additionally, surveys excel in raw performance compared to our temperature model because it collects absolute poverty statistics directly from households, without the added noise created by machine learning algorithms. However, the demonstrated use of tracking regional poverty trends proves that the temperature model approaches survey data's poverty tracking performance Furthermore, the model is reproducible because the only data needed is temperature data and satellite imagery, widely available through organizations such as NOAA or Google. The model tested in this study is a simpler and accurate alternative to the expensive surveys of today.

Given the shortcomings of a temperature model, we believe that the model performance may be increased through data collection improvements. While spatial average was conducted via a clustering system, temporal averaging can also be conducted by utilizing more data. WorldClim's averaged monthly datasets can be used to limit the effect of climatic anomalies on the model. Thus, the CNN model would train on an intra-annual subset of data. Additionally, the model's three data sources - satellite data, temperature data, and consumption data - have gaps in their collection time periods. Satellite data is the most recent, taken within 1 year, while temperature data and consumption data are from the same time periods. Shifting to the use of Google Earth Engine would allow us to control the time parameter of image queries to better align with the temperature and consumption data. These changes would increase the performance of our model when regressing.

In addition to altering the data collection methods, expanding the study to additional countries would allow us to test the temperature model's generalizability. In this study we prefer generalizing, both in CNN training and ridge training, as a way to reduce overfitting in Ethiopia's own distinct climate and procuring unbiased model results. When including other countries in a future study, an additional step would need to be taken to ensure that a variety of economic regions and climates are being counted to avoid a strict adherence to the high-temperature high-poverty scenario. There may be countries at risk from climate change but with resources available to actively fight against it or countries not affected by climate change but instead other poverty determinants such as access to education, healthcare, and a robust political system. Categorizing these scenarios and representing them through inclusive datasets is critical to extrapolating the predictions of the temperature model.

The ubiquitous nature of data collected actively and passively has the potential to build a greater understanding of demographics in countries where wealth and consumption data is lacking. However, the same data sources may fuel a struggle for data sovereignty and ethical implications. This warrants a minor discussion of the implications of global, instantaneous, data-driven poverty predictions.  In this study, poverty is viewed through the lens of annual consumption which is a predominantly capital-driven view of community status. In certain communities, a welfare ranking may function as a more inclusive framework for analyzing poverty. Welfare-based studies also take into consideration agricultural communities that practice subsistence farming, growing only as much food needed to support their household. To properly acknowledge the various types of global lifestyles, it is important to present poverty and consumption predictions within the historical context of a community.

Using a transfer learning model utilizing high resolution remote sensing data and global average temperature data, we accurately predict poverty in Ethiopia. A convolutional neural network based on the advanced VGG-11 standard is trained on temperature data to intrinsically detect features in satellite images that could be useful for predicting poverty. These features are aggregated to produce fully-fledged poverty estimates comparable - and in some cases even exceeding - in accuracy to other established poverty tracking methods. This novel method cuts down on crucial time needed to collect survey data and enables a more efficient process for maintaining poverty standards in developing nations. This technology has the full potential to revolutionize how countries and organizations work across the globe to address poverty.

## References

[1] UN Data Revolution Group, "A World That Counts: Mobilizing The Data Revolution for Sustainable Development," New York, USA. Report. 2014.

[2] P. Lanjouw, and N. Yoshida, "Poverty Monitoring Under Acute Data Constraints: A Role for Imputation Methods?" Inequality Matters, Nov. 2021.

[3] G. Stecklov, and A. Weinreb, "Improving the Quality of Data and Impact-Evolution Studies in Developing Countries," Inter-American Development Bank, Washington, D.C., USA. Report. IDB-TN-123, pp. 34-53, May 2010.

[4] X. Michael, J. Neal, B. Marshall, D. Lobell, and S. Ermon, "Transfer Learning from Deep Features for Remote Sensing and Poverty Mapping," 30th AAAI Conference on Artificial Intelligence, 2016.

[5] C. Yeh, A. Perez, A. Driscoll, et al, "Using publicly available satellite imagery and deep learning to understand economic well-being in Africa," Nature Communications 11, 2583, 2020  DOI: 10.1038/s41467-020-16185-w

[6] UN Department of Economic and Social Affairs, "World Social Report: Inequality in a Rapidly Changing World," New York, USA. Report. 2020.

[7] T. W. Hertel, and S. D. Rosch, "Climate Change, Agriculture and Poverty," AAEA Applied Economics Perspectives and Policies Journal, pp. 4-7, November 2010.

[8] M. Dell, B.F. Jones, and B.A. Olken, " Climate Change and Economic Growth: Evidence from the Last Half Century," National Bureau of Economic Research, June 2008. DOI: 10.3386/w14132

[9] M. Prange, T. Wilke, and F. P. Wesselingh, "The other side of sea level change," Commun Earth Environ vol. 1, no. 69, December 2020.

[10] A. Hashemi, H. Cruikshank, A. Cheshmehzangi, "Improving Thermal Comfort in Low-income Tropical Housing: The Case of Uganda," presented at ZEMCH 2015 International Conference, Lecce Italy, 2015.

[11] X. Han et al, "Pre-trained models: Past, present and future," AI Open, vol. 2, pp.. 225-250, 2021.

[12] A. O. Salau and S. Jain, "Feature Extraction: A Survey of the Types, Techniques, Applications," 2019 International Conference on Signal Processing and Communication (ICSC), 2019, pp. 158-164.

[13] A.E. Hoerl, and R.W. Kennard, "Ridge Regression: Biased Estimation for Nonorthogonal Problems," Technometrics, vol. 12, no. 1, pp. 55–67, 1970.

[14] D. Berrar, "Cross-Validation," in Encyclopedia of Bioinformatics and Computational Biology, Cambridge, MA: Academic Press, 2019, pp. 542-545.

[15] A. Deaton, "Household Surveys, Consumption, and the Measurement of Poverty," Economic Systems Research, vol. 15, no. 2, pp. 135-159, 2003

[16] N. Ritter, and M. Ruth, "The GeoTiff data interchange standard for raster geographic images," International Journal of Remote Sensing, vol. 18, no. 7, pp. 1637-1647, 1997.

[17] J. Wu, and H. Hamdan, "Parsimonious Gaussian mixture models of general family for binned data clustering: Mixture approach," 2012 IEEE 10th International Symposium on Applied Machine Intelligence and Informatics (SAMI), 2012, pp. 283-288.

[18] K. Simonyan, and A. Zisserman, "Very deep convolutional networks for large-scale image recognition," arXiv preprint, 2014.

[19] L. Perez, and J. Wang, "The Effectiveness of Data Augmentation in Image Classification using Deep Learning," arXiv preprint, 2017.

[20] B. Tang, Y. Liu, Y. Sun, and D.S. Matteson, "Predicting poverty with vegetation index." AAEA Applied Economic Perspectives and Policies Journal, pp. 1-16, 2021.

## Competing Interests
We declare no competing interests.

## Appendix: Code Availability
Code for this research was adapted from the pythonification work of Jean et al. by Jatin Mathur to examine the temperature-consumption relationship. All work is available at:
https://github.com/omshah2006/Predicting-Poverty-Ethiopia